\documentclass[10pt,journal,compsoc]{IEEEtran}
\usepackage[american]{babel}

\usepackage{graphicx}
\usepackage{amsmath}
\usepackage{amssymb}
\usepackage{mathrsfs}

\usepackage{overpic}
\usepackage{multirow}
\usepackage{bm}
\usepackage{booktabs}
\usepackage[ruled,linesnumbered]{algorithm2e}
\usepackage{subcaption}
\usepackage{lipsum}
\usepackage{colortbl}
\definecolor{mygray}{gray}{.8}

\usepackage[pagebackref=false,breaklinks=false,letterpaper=false,colorlinks,bookmarks=false]{hyperref}

\def\ie{\emph{i.e.}}
\def\eg{\emph{e.g.}}

\newcommand{\rev}[1]{\textcolor{red}{#1}}

\newcommand{\blue}[1]{\textcolor{black}{#1}}

\usepackage{wrapfig}
\usepackage{tikz,pgfplots}

\usepackage{ragged2e}
\renewcommand{\raggedright}{\leftskip=0pt \rightskip=0pt plus 0cm}

\ifCLASSOPTIONcompsoc
  \usepackage[nocompress]{cite}
\else
  \usepackage{cite}
\fi

\begin{document}

\title{Anchor-driven Multi-modal Multi-scale Expert Selection for Survival Prediction}

 \author{Tao Zhou,~\IEEEmembership{Senior~Member,~IEEE,}
    Ying Hu, 
 	Huazhu Fu,~\IEEEmembership{Senior~Member,~IEEE,}
 	Yi Zhou~\IEEEmembership{Senior~Member,~IEEE,}
 	Xiao-Jun Wu,
 	Haibin~Ling, ~\IEEEmembership{Fellow,~IEEE},\\

\IEEEcompsocitemizethanks{
\IEEEcompsocthanksitem T. Zhou and Y.~Hu are with the School of Computer Science and Engineering, Nanjing University of Science and Technology, Nanjing 210094, China.
\IEEEcompsocthanksitem H.~Fu is with s with the Institute of Advanced Intelligence and Computing (IAIC), Agency for Science, Technology and Research (A*STAR), Singapore 138632. 
\IEEEcompsocthanksitem Y. Zhou is with the School of Computer Science and Engineering, Southeast University, Nanjing 211189, China. 
\IEEEcompsocthanksitem W.-J. Wu is with the School of Artificial Intelligence and Computer Science, Jiangnan University, Wuxi 214122, China.
\IEEEcompsocthanksitem H. Ling is with the Westlake Intelligent Computing and Application Lab, Dept of Artificial Intelligence, Westlake University, Hangzhou, China. 
}
}

\markboth{AM2ES for Survival Prediction}%
{Shell \MakeLowercase{\textit{et al.}}: Bare Demo of IEEEtran.cls for Computer Society Journals}

\IEEEtitleabstractindextext{
\begin{abstract}

\raggedright{
The integrative analysis of histopathological Whole-Slide Images (WSIs) and transcriptomic profiles holds significant promise for cancer survival prediction. However, existing methods typically project multi-modal features directly into a shared latent space without explicit alignment, leading to the entanglement of mismatched morphological cues and molecular signals. Furthermore, current fusion strategies often treat the extreme spatial heterogeneity of WSIs uniformly, lacking mechanisms to adaptively prioritize clinically relevant tissue scales for individual patients. To address these limitations, we propose an Anchor-driven Multi-modal Multi-scale Expert Selection (AM$^2$ES) framework for survival prediction. Specifically, we present an Anchor-driven Multi-modal Fusion (AMF) module, which introduces learnable semantic anchors as cross-modal mediators to bridge the semantic gap by enforcing a structurally regularized alignment between transcriptomic features and multi-scale pathology representations. Built upon this aligned semantic space, we further design a Hierarchical Mixture-of-Experts (H-MoE) selection module to decouple the hierarchical prognostic selection process. Mimicking the pathologist's diagnostic workflow, H-MoE performs (i) Intra-scale Expert Filtering to discriminatively identify salient tumor regions within each magnification, and (ii) Inter-scale Hierarchy Routing to dynamically weight and select the most informative resolution levels. Extensive experiments on multiple TCGA cancer cohorts demonstrate that our AM$^2$ES achieves state-of-the-art performance while offering fine-grained interpretability by visualizing how specific molecular pathways drive the expert routing decisions across tissue scales. The code will be released at \href{https://github.com/taozh2017/AM2ES}{https://github.com/taozh2017/AM2ES}
}.

\end{abstract}

}

\maketitle

\IEEEdisplaynontitleabstractindextext

\IEEEpeerreviewmaketitle

\section{Introduction}\label{sec:introduction}

\IEEEPARstart{A}{ccurate} survival prediction is a cornerstone of precision oncology. To tackle the challenges posed by the heterogeneity of cancer prognosis, researchers have drawn inspiration from clinical diagnostic workflows and focused on multimodal learning~\cite{lipkova2022artificial}. In clinical practice, pathologists typically integrate fine-grained spatial context from Whole Slide Images (WSIs) with molecular profiles to determine risk stratification. Translating this workflow into computational models, researchers have introduced various WSI-Omics fusion frameworks~\cite{chen2021multimodal, han2025inter,li2021dual,di2022generating,yan2025pathway}, enabling the extraction of complementary phenotypic and molecular determinants of patient outcomes from high-dimensional biomedical data.

\begin{figure}[!t]
	\centering
    \includegraphics[width=0.5\textwidth]{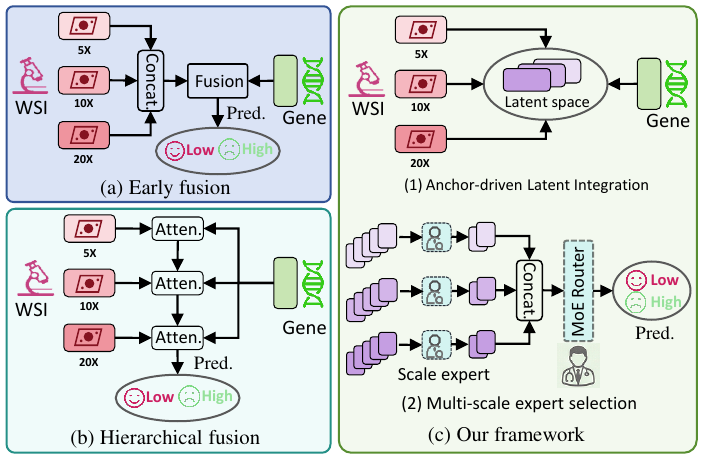}\vspace{-0.15cm}
      \vspace{-0.2cm}
	\caption{\footnotesize Comparison of multimodal fusion paradigms in WSI-Omics survival analysis: (a) Early Fusion, which concatenates multi-scale pathological features prior to cross-modal interaction; (b) Hierarchical Fusion, which sequentially integrates multi-scale features layer by layer in a fixed cascading manner; and (c) Our proposed AM$^2$ES framework, which explicitly decouples semantic alignment via dynamic anchors from hierarchical scale selection via H-MoE.}
    \label{fig:motivation}\vspace{-0.25cm}
\end{figure}

Despite these significant advancements, existing WSI-Omics frameworks face intertwined challenges in both cross-modal alignment and spatial hierarchy modeling (as overviewed in Fig.~\ref{fig:motivation}). First, bridging the severe semantic gap between pixel-level morphology and abstract gene expression vectors remains a highly underconstrained problem. Yet, current models typically rely on standard global cross-attention to fuse these disparate modalities in a single step~\cite{MCAT_2021_ICCV, MoCAT_2023_ICCV}. This ``flat" interaction inevitably entangles two distinct objectives: semantic alignment and prognostic selection. By forcing gigapixel WSIs and 1D molecular signatures into a shared space simultaneously, the network is compelled to learn what to align (matching tissue patches to biological pathways) and what to select (identifying features that drive survival) all at once. Without explicit decoupling, this entanglement frequently leads to modality overfitting and feature collapse. Consequently, attention maps often highlight spurious background noise rather than meaningful biomarkers, severely compromising clinical interpretability.

Additionally, existing methods struggle to adaptively capture the clinical hierarchy of pathological features. In diagnostic pathology, survival-relevant patterns manifest at distinct anatomical scales. For instance, macro-tissue architecture and tumor margins are evaluated at low magnification (\eg, $5\times$), while micro-cellular atypia and mitotic figures require high magnification (\eg, $20\times$). However, existing fusion architectures struggle to adaptively capture this clinical hierarchy, generally falling into two paradigms (Fig.~\ref{fig:motivation}(a-b)): early fusion and hierarchical Fusion. Early fusion methods (Fig.~\ref{fig:motivation}(a)) typically extract multi-scale WSI features using conventional flattening strategies~\cite{Hashimoto_2020_CVPR, DSMIL_2021_CVPR} and concatenate them into a unified sequence before interaction. This strategy treats distinct biological structures as equivalent visual tokens, inevitably leading to scale confusion, noise propagation, and the dilution of scale-specific prognostic signals. Conversely, Hierarchical fusion methods (Fig.~\ref{fig:motivation}(b)) adopt a sequential pipeline, integrating different magnifications layer by layer in a fixed cascading manner~\cite{pati2022hierarchical, HiMT_2022, UMSA_2025}. By assuming a fixed order of feature importance across all patients, these models fail to adaptively prioritize the specific resolutions that are most relevant to prognosis for individual heterogeneous cases.

To address the dual challenges of entangled alignment and architectural rigidity, an effective multimodal framework must explicitly decouple these processes. In this paper, we propose AM$^2$ES, a novel Anchor-driven Multi-modal Multi-scale Expert Selection framework for cancer survival prediction using multi-modal data. As illustrated in Fig.~\ref{fig:motivation}(c), our model effectively bridges the gap between WSIs and transcriptomic profiles while selecting informative features to enhance prediction performance. 
Specifically, to resolve the modality gap, we present an Anchor-driven Multi-modal Fusion (AMF) module. Rather than relying on static prototypes, our anchors are dynamically conditioned on global patient contexts, serving as personalized semantic mediators. Through structural regularization, these anchors explicitly align transcriptomic and pathological representations, ensuring that the extracted visual features are biologically grounded in the corresponding molecular profiles. 
To address the scale hierarchy, we develop a Hierarchical Mixture-of-Experts (H-MoE) selection strategy. This strategy operates in two stages: (1) an Intra-scale Filtering stage, which employs scoring experts to rigorously isolate salient multi-modal representations and filter out background noise within each distinct resolution; and (2) an Inter-scale Fusion stage, which leverages rank-based alignment and feature-level MoE to dynamically and efficiently recalibrate and synthesize evidence across different spatial scales. This joint design enables the model to adaptively weigh multi-scale evidence tailored to each patient’s unique pathology for accurate survival prediction.

In summary, our contributions are four-fold:
\begin{itemize}
    \item We propose AM$^2$ES, a novel framework that explicitly decouples cross-modal semantic alignment from hierarchical prognostic selection, addressing the entanglement and flatness issues prevalent in existing multimodal survival models.
    
    \item We present a dynamic anchor-driven multi-modal fusion module where learnable and patient-contextualized anchors serve as semantic bridges. By enforcing orthogonality and sparsity constraints, this module ensures that pathological features are diverse, stable, and biologically grounded in their corresponding transcriptomic profiles.
    
    \item We propose a two-stage H-MoE mechanism for expert selection. It integrates intra-scale gating for noise filtration and inter-scale dynamic recalibration for adaptive resolution fusion, achieving a balance between multi-scale expressiveness and stable routing. 
    
    \item Extensive experiments on multiple TCGA cohorts demonstrate that our AM$^2$ES achieves state-of-the-art performance. Crucially, our framework offers fine-grained interpretability by visually tracing both the anchor-driven semantic shifts and the patient-specific scale preferences.

\end{itemize}

\begin{figure*}[!t]
	\centering
	\footnotesize
	\begin{overpic}[width=0.98\textwidth]{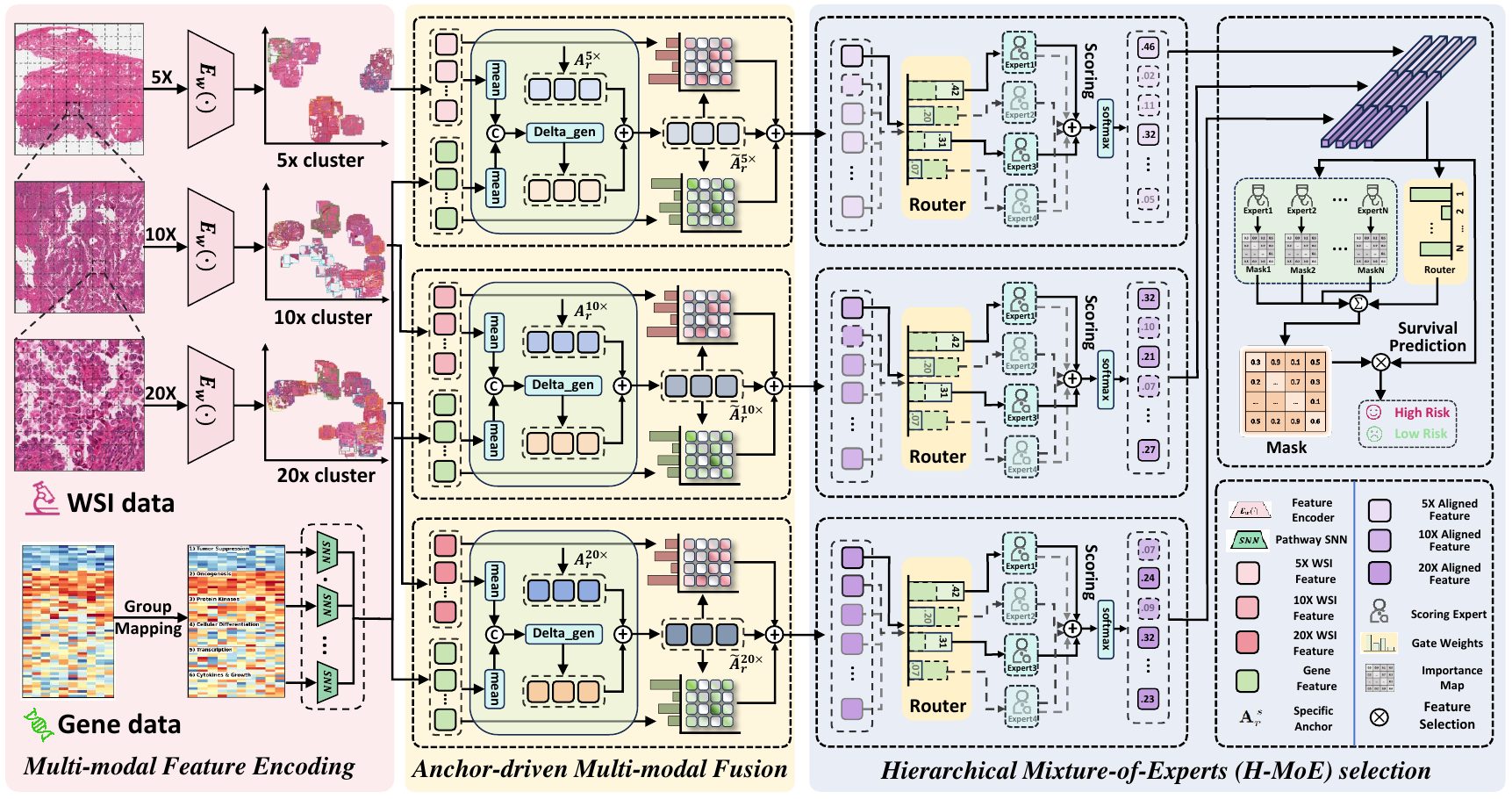}
	\end{overpic} \vspace{-0.1cm}
	\caption{\footnotesize 
    Overview of the proposed AM$^2$ES framework for multi-modal survival prediction. 
    The pipeline integrates multi-scale WSIs and transcriptomic profiles through two synergistic stages: 
    (1) Dynamic Anchor-Driven Alignment: Patient-specific anchors are generated conditioned on global genomic and morphological contexts. These anchors serve as semantic mediators to align multi-modal features, where structural regularization (orthogonality and sparsity constraints) is enforced to capture diverse biological patterns. 
    (2) Hierarchical Mixture-of-Experts (H-MoE): A two-stage mechanism handles tissue heterogeneity. The Intra-scale Expert Selection module evaluates and filters morphological signals within each resolution (\emph{i.e.}, $5\times, 10\times, 20\times$) to remove redundancy. Subsequently, the Inter-scale MoE Fusion module employs rank-based alignment and feature-level gating to dynamically recalibrate and aggregate prognostic evidence across scales for final survival estimation.
    }
	\label{fig:Overview}\vspace{-0.15cm}
\end{figure*}

\section{Related Work}\label{sec:Related_works} 

\subsection{Multimodal Learning for Cancer Prognosis}

Integrating whole-slide histopathology with transcriptomic data has become a pivotal direction for enhancing cancer prognostication. Early multimodal approaches~\cite{nikolaou2025machine,fan2023pancancer} predominantly fused WSI-derived features with genomic signatures within a unified embedding space via simple concatenation, aiming to capture complementary cues. Mobadersany et al.~\cite{mobadersany2018predicting} pioneered the joint modeling of histology and gene expression using convolutional networks. Subsequent works, such as Pathomic Fusion~\cite{chen2020pathomic}, incorporated tensor-based gating to model modality interactions. More recently, Transformer-based frameworks have dominated the field. Approaches like MCAT~\cite{chen2021multimodal} and MOTCat~\cite{xu2023multimodal} utilize co-attention mechanisms to align visual patches with genomic embeddings. While effective, these methods typically intertwine representation learning and prognostic fusion within a single ``flat" interaction step. They lack explicit semantic mediators, forcing the network to implicitly learn cross-modal correspondence while simultaneously performing risk prediction. This often results in modality overfitting and feature entanglement, directly motivating the explicitly decoupled anchor-driven multi-modal fusion paradigm in our framework.

\subsection{Multi-scale Representation in Computational Pathology}

Histopathology images exhibit substantial heterogeneity across scales, where biological signals are distributed hierarchically: global tissue phenotypes (\eg, tumor-stroma ratio) are visible at low magnification ($5\times$), while fine-grained cellular atypia requires high magnification ($20\times$). Consequently, multi-scale learning has become essential for comprehensive WSI analysis~\cite{paverd2024radiology,wang2021hierarchical,wu2024leveraging}. Early approaches extended Multiple Instance Learning (MIL) to handle multi-resolution inputs. For instance, Ilse et al.~\cite{ilse2018attention} proposed attention-based pooling to weigh instances, while DS-MIL~\cite{li2021dual} introduced a dual-stream architecture that utilizes low-magnification scores to guide the selection of high-magnification patches. To better capture spatial context, Graph Convolutional Networks (GCNs) such as Patch-GCN~\cite{chen2021whole} have been employed to model the topological relationships between multi-scale tissue regions.

More recently, Transformer-based architectures have advanced WSI analysis by modeling long-range dependencies. TransMIL~\cite{shao2021transmil} employs pyramid-like token aggregation, whereas HIPT~\cite{chen2022scaling} uses a hierarchical vision transformer to capture features from cellular to regional levels. Prompt-based methods, such as MSCPT~\cite{han2025mscpt}, further leverage learnable prompts to adaptively capture multi-scale context for tasks such as WSI classification. Despite these advances, {most existing methods combine multi-scale features through fixed strategies, such as early or hierarchical fusion. They lack explicit mechanisms to identify the prognostically informative scale for each \textit{individual} patient}. 



\subsection{Anchor-driven Prototypes and Multi-modal Fusion}

Prototype- and anchor-based learning has emerged as an effective paradigm for structured representation learning~\cite{ning2023madav2,ming2021dynamic}. In computer vision, DETR~\cite{carion2020end} introduced learned query embeddings as semantic extractors. This concept has been extended to general multi-modal learning by recent frameworks~\cite{jeong2024anchors}, which utilized anchors to construct an optimal unified representation space for heterogeneous modalities. However, existing anchor-based approaches in survival analysis typically treat anchors as fixed parameters. While methods like DPU~\cite{li2025dpu} have introduced dynamic prototype updating to handle out-of-distribution shifts, such dynamic mechanisms have rarely been adapted for patient-specific cross-modal fusion in survival tasks. Most current models seldom adapt anchors based on the joint transcriptomic--histological context. In contrast, the Anchor-driven Multi-modal Fusion (AMF) module in our AM$^2$ES introduces dynamic, patient-contextualized anchors. These anchors act as personalized semantic bridges, conditioned on global RNA profiles, to explicitly probe and align morphological features via structural regularization.

\subsection{Mixture-of-Experts Selection Mechanisms}

Mixture-of-Experts (MoE)~\cite{shazeer2017outrageously} has re-emerged as a powerful paradigm for conditional computation, where sparse gates dynamically activate specialized expert units. While modern variants like Switch Transformers~\cite{fedus2022switch} and V-MoE~\cite{riquelme2021scaling} have demonstrated scalability in large-scale language and vision models, their adoption in computational pathology remains nascent. In medical imaging, MoE mechanisms have primarily been explored to handle domain variability and stain heterogeneity~\cite{li2024generalizable, jiang2024med}. However, these applications generally adopt single-stage structures without explicit semantic disentanglement. More critically, existing WSI–omics survival models rarely combine MoE gating with explicit multi-modal fusion strategies. Most frameworks fuse modalities early or late in a shared latent space, lacking mechanisms to perform scale-aware or survival-aware expert selection. 

\section{Proposed Method}

\subsection{Problem Definition}\label{sec:problem_def}

We consider a survival prediction task on a cohort of $N$ patients, denoted as $\mathcal{D} = \{(X_{i}, t_{i}, c_{i})\}_{i=1}^{N}$. For the $i$-th patient, the multi-modal input is defined as $X_{i} = \{X_{i}^{p}, X_{i}^{g}\}$, where $X_{i}^{p}$ represents a multi-scale set of pathological whole-slide images (WSIs), and $X_{i}^{g}$ denotes the genomic profile. Let $T_i$ be the continuous random variable representing the true time-to-event (survival time) for the $i$-th patient. The survival outcome is composed of the observed time $t_{i} \in \mathbb{R}^{+}$ and the censorship status $c_{i} \in \{0, 1\}$. In our task, $c_{i} = 1$ denotes a right-censored observation, while $c_{i} = 0$ indicates an uncensored event (death). 

To effectively model the survival distribution, we adopt the discrete-time survival paradigm. We partition the continuous time horizon into $R$ non-overlapping intervals: $[q_{0}, q_{1}), [q_{1}, q_{2}), \ldots, [q_{R-1}, q_{R})$. The bin boundaries $\{q_j\}_{j=1}^{R}$ are determined by the $R$-quantiles of survival times from the uncensored subset, with $q_0 = \min_i(t_i) - \epsilon$ and $q_R = \max_i(t_i) + \epsilon$ to cover the entire cohort. The goal is to estimate the discrete hazard probability $p_{i,j}$:
\begin{equation}
    p_{i,j} = \Pr(T_{i} \in [q_{j-1}, q_{j}) \mid T_{i} \ge q_{j-1}),
\end{equation}
which represents the conditional probability of the event occurring in the $j$-th interval, given that the patient survived until the start of that interval. The survival function at the $k$-th time interval is subsequently derived as follows:
\begin{equation}
    \label{eq:survival_function}
    S_{i}(k) = \prod_{j=1}^{k} (1 - p_{i,j}), \quad \text{with } S_{i}(0) = 1.
\end{equation}

\subsection{Overview of Architecture}


{In this work, we propose AM$^2$ES for survival prediction from multi-scale WSI and transcriptomic data (Fig.~\ref{fig:Overview}). The framework comprises three components: multi-modal feature encoding, Anchor-driven Multi-modal Fusion (AMF), and Hierarchical Mixture-of-Experts (H-MoE) selection. First, independent encoders extract multi-scale WSI features ($5\times, 10\times, 20\times$) and RNA embeddings, while a Transformer-based clustering module summarizes WSI features into semantic prototypes. Next, AMF employs learnable anchors to align image prototypes and genomic embeddings in a shared representation space. Finally, H-MoE selects informative features within each scale and adaptively integrates them across scales for survival prediction.}

\subsection{Multi-scale Histopathology and Genomic Feature Encoding}

{We employ separate encoding streams to derive multi-scale morphological representations from WSIs and biologically grouped descriptors from genomic profiles. The obtained image features and genomic embeddings provide structured inputs for the subsequent anchor-driven fusion module.} {Moreover, rather than prematurely blending representations across resolutions through heuristic pooling, our framework hierarchically integrates multiscale features.}

\subsubsection{Multi-scale WSI Representation Learning}\label{sec:wsi_learning}

{For patient $i$, the WSI at scale $s \in \{5\times, 10\times, 20\times\}$ is represented as a bag of patches
$W_i^s=\{w_{i,n}^s\}_{n=1}^{M_i^s}$.
Each patch is encoded by a shared visual encoder $E_w(\cdot)$ and projected into a $D$-dimensional latent space:}
\begin{equation}
\mathbf{v}^s_{i,n} = \phi\big(E_w(\mathbf{w}^s_{i,n})\big) \in \mathbb{R}^{D},
\end{equation}
where $\phi(\cdot)$ denotes the projection head, implemented as a fully connected layer with ReLU activation.

Then, we employ a Transformer-based clustering module to aggregate the dense patch embeddings into scale-specific $P^s$ representative semantic prototypes. {Specifically, we initialize a set of learnable prototype queries $\mathbf{Q}^s \in \mathbb{R}^{P^s \times D}$, which serve as learnable cluster centers in the embedding space. Unlike traditional clustering algorithms that rely on static distance metrics, our module utilizes the cross-attention mechanism to perform task-driven soft-assignment. Given the input dense patch embeddings $\mathcal{V}_i^s \in \mathbb{R}^{M_i^s \times D}$ (serving as both keys and values), the semantic prototypes $\mathcal{H}_i^s$ are computed via standard Multi-Head Cross-Attention (MHCA) followed by a Feed-Forward Network (FFN):
\begin{equation}
\mathcal{H}_i^s = \mathrm{FFN}(\mathrm{MHCA}(\mathbf{Q}^s, \mathcal{V}_i^s, \mathcal{V}_i^s)) \in \mathbb{R}^{P^s \times D},
\end{equation}
where the cross-attention mechanism implicitly calculates the normalized contributions of all constituent patches to each specific prototype.
By optimizing the model end-to-end, $\mathbf{Q}^s$ automatically converges to representative centers that capture distinct morphological patterns without requiring heuristic clustering steps. This attention-driven soft-assignment dynamically groups morphologically similar patches into the patient-level WSI representation $\mathcal{H}_i^s=\left\{\mathbf{h}^s_{i,n}\right\}_{n=1}^{P^s}$. These multi-scale embeddings efficiently preserve coarse-to-fine structural information while strictly controlling computational complexity for downstream fusion.}


\subsubsection{Biologically-grouped Genomic Encoding}\label{sec:genomic_learning}

{We partition the genomic profile of the $i$-th patient into $M_g$ biologically defined groups}, \ie, $\mathbf{X}_i^g = \{\mathbf{x}_{i,1}^g,\ldots, \mathbf{x}_{i,M_g}^g\}$, where each subset $\mathbf{x}_{i,j}^g \in \mathbb{R}^{d_j}$ contains $d_j$ genes associated with a specific biological process. To accommodate these differences in dimensionality and statistical properties, we employ category-specific encoders. Then, each genomic subset is projected into a shared latent dimension $D$ via a lightweight network $f_j(\cdot)$ as follows: 
\begin{equation}
    \label{eq:genomic_encoding}
    \mathbf{g}_{i,j} = f_j(\mathbf{x}_{i,j}^g) \in \mathbb{R}^{D}.
\end{equation}

{The resulting genomic bag is defined as $\mathbf{G}_i = \left\{ \mathbf{g}_{i,1}, \ldots, \mathbf{g}_{i,M_g} \right\} \in \mathbb{R}^{M_g \times D}$, which preserves pathway-level organization for anchor-driven fusion.}


\begin{figure}[!t]
	\centering
    \includegraphics[width=0.49\textwidth]{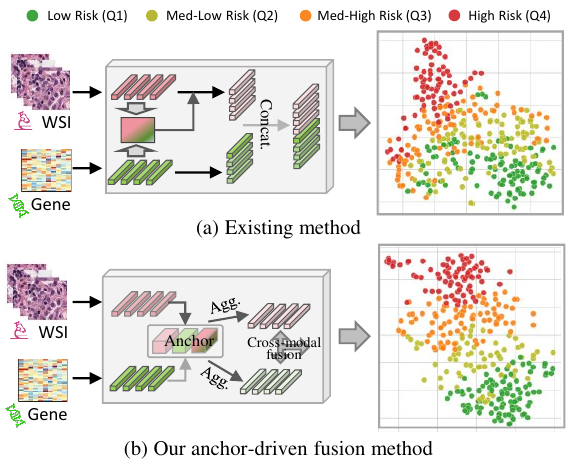}\vspace{-0.15cm}
      \vspace{-0.2cm}
	\caption{\footnotesize Comparison of multi-modal fusion paradigms in WSI-Omics survival analysis. (a) Existing methods~\cite{xu2023multimodal} leverage co-attention and direct concatenation to fuse cross-modal data. (b) Our anchor-driven fusion strategy acts as a semantic bridge, explicitly aligning cross-modal features via learnable anchors to construct a highly discriminative and structured representation space.}
    \label{fig:motivation_anchor}\vspace{-0.25cm}
\end{figure}

\subsection{Anchor-driven Multi-modal Fusion}


In multi-modal survival prediction, histology and transcriptomics provide complementary yet distinct biological information. As illustrated in Fig.~\ref{fig:motivation_anchor}(a), {existing methods~\cite{xu2023multimodal} typically rely on co-attention and direct concatenation to fuse cross-modal data, which often leads to an entangled latent space where patients across different risk levels are not well separated}. To address this limitation, {we introduce learnable, scale-specific cross-modal anchors. Inspired by prototype learning, these anchors act as semantic bridges that align hierarchical tissue representations with gene-expression embeddings at each biological scale. Unlike static queries, the anchors are dynamic and patient-conditioned. Through bidirectional attention, they establish structured correspondences between high-level phenotypic patterns and underlying molecular pathway activities. As illustrated by the t-SNE visualization in Fig.~\ref{fig:motivation_anchor}(b), anchor-driven integration yields a more discriminative latent space and clearly separates distinct survival risk profiles.}


\subsubsection{Dynamic Anchor Generation}

{For each scale $s$, we maintain a set of $K$ learnable reference anchors
$A_r^s \in \mathbb{R}^{K \times D}$. Patient-specific anchors are obtained by conditioning these reference anchors on the global genomic and scale-specific morphological contexts.} 
To adapt these anchors to individual patients, we compute global context vectors from the encoded morphological features $\mathcal{H}_i^s$ and $\mathbf{G}_i$ (detailed in Sec.~\ref{sec:wsi_learning} and \ref{sec:genomic_learning}). More concretely, the global genomic context $\mathbf{c}^{(g)}_i$ and the scale-specific morphological context $\mathbf{c}^{(w,s)}_i$ are derived via average pooling as follows:
\begin{equation}
    \mathbf{c}^{(g)}_i = \frac{1}{M_g} \sum_{m=1}^{M_g} \mathbf{g}_{i,m}, \quad
    \mathbf{c}^{(w,s)}_i = \frac{1}{P^s} \sum_{n=1}^{P^s} \mathbf{h}^{s}_{i,n}.
\end{equation}

We then generate patient-specific dynamic anchors $\widetilde{\mathbf{A}}_i^{\,s}$ by injecting these global signals into the reference anchors:
\begin{equation}
    \widetilde{\mathbf{A}}_i^{\,s} = \mathbf{A}_{r}^{s} + f_{\mathrm{cond}}^{s} [\mathbf{c}^{(g)}_i \parallel \mathbf{c}^{(w,s)}_i ],
\end{equation}
where $[ \cdot \parallel \cdot ]$ denotes concatenation, and $f_{\mathrm{cond}}^{s}: \mathbb{R}^{2D} \to \mathbb{R}^{K \times D}$ is a dynamic calibration network. 
This mechanism ensures that the anchors $\widetilde{\mathbf{A}}_i^{\,s}$ are tailored to the specific tumor microenvironment of the $i$-th patient at resolution $s$.

\subsubsection{Dual-stream Anchor Attention}

Using the patient-conditioned anchors $\widetilde{\mathbf{A}}_i^s$, we perform structured multi-modal alignment in parallel for each scale. Each anchor acts as a dual query to retrieve evidence from both histology and genomics. {For a given scale $s$ and its $k$-th anchor $\widetilde{\mathbf{a}}^{\,s}_{i,k}$, the aggregated morphological evidence $\mathbf{z}^{s,\text{WSI}}_{i,k} \in \mathbb{R}^D$ is computed by attending over the morphological prototypes $\mathcal{H}_i^{\,s}$ as follows:
\begin{equation}
    \mathbf{z}^{s,\text{WSI}}_{i,k} = \sum_{n=1}^{P^s} \underset{n}{\mathrm{softmax}} \left( \frac{(\mathbf{W}^{Q}_w \widetilde{\mathbf{a}}^{\,s}_{i,k})^\top (\mathbf{W}^K_w \mathbf{h}^{s}_{i,n})}{\sqrt{D}} \right) (\mathbf{W}^V_w \mathbf{h}^{s}_{i,n}),
\end{equation}
where $\mathbf{W}^{Q}_w, \mathbf{W}^K_w, \mathbf{W}^V_w \in \mathbb{R}^{D \times D}$ are learnable projection matrices. The softmax is applied over all prototypes $n$, ensuring selective attention to scale-specific morphology patterns. 
}

{Similarly, the same anchor queries the genomic bag $\mathbf{G}_i$ to retrieve molecular evidence $\mathbf{z}^{s,\text{GEN}}_{i,k} \in \mathbb{R}^D$ via a structurally symmetric operation:
\begin{equation}
\mathbf{z}^{s,\text{GEN}}_{i,k} = \sum_{m=1}^{M_g} \underset{m}{\mathrm{softmax}} \left( \frac{(\mathbf{W}^{Q}_g \widetilde{\mathbf{a}}^{\,s}_{i,k})^\top (\mathbf{W}^K_g \mathbf{g}_{i,m})}{\sqrt{D}} \right) (\mathbf{W}^V_g \mathbf{g}_{i,m}),
\end{equation}
where $\mathbf{W}^{Q}_g, \mathbf{W}^K_g, \mathbf{W}^V_g \in \mathbb{R}^{D \times D}$ are learnable genomic projection matrices.}

To synthesize the retrieved cross-modal evidence while preserving the intrinsic semantics of the anchors, we employ a residual fusion mechanism: 
\begin{equation}
    \mathbf{u}_{i,k}^s = \mathrm{LN} \left( \sigma (\mathbf{W}_f [\mathbf{z}^{s,\text{WSI}}_{i,k} \parallel  \mathbf{z}^{s,\text{GEN}}_{i,k}] 
      + \mathbf{b}_f )+ \widetilde{\mathbf{a}}^{\,s}_{i,k} \right),
\end{equation}
where $\mathbf{W}_f \in \mathbb{R}^{D \times 2D}$ and $\mathbf{b}_f \in \mathbb{R}^D$ are the fusion layer parameters, $\sigma(\cdot)$ denotes the ReLU activation, and $\text{LN}(\cdot)$ represents Layer Normalization. This process yields a collection of aligned cross-modal features $\mathbb{U}_i = \{ \mathcal{U}_i^{5\times}, \mathcal{U}_i^{10\times}, \mathcal{U}_i^{20\times} \}$, where each scale-specific set $\mathcal{U}_i^s = \{ \mathbf{u}_{i,k}^s \}_{k=1}^K$ contains $K$ refined multi-modal candidates. The distinct feature sets in $\mathbb{U}_i$ are preserved to allow for dynamic and context-aware selection both within and across scales in the subsequent process. 

\subsubsection{Regularization for Stable Anchor Fusion}\label{sec:anchor_reg}

We regularize the patient-specific anchors to encourage diversity and sparsity. For each scale $s$, an orthogonality term promotes distinct anchor directions, while an $\ell_1$ penalty encourages compact anchor activations. 
Let $\widetilde{\mathbf{A}}_i^{\,s} \in \mathbb{R}^{K \times D}$ denote the matrix of patient-specific anchors at scale $s$, where each row $\widetilde{\mathbf{a}}_{i,k}^{\,s}$ represents a distinct latent expert. We first enforce an orthogonality constraint to encourage diversity among the $K$ anchors. By penalizing the correlation between different anchors, we ensure they span distinct semantic directions in the latent space by
\begin{equation}
    \mathcal{L}_{\mathrm{orth}}^{s} = \left\| \widehat{\mathbf{A}}_i^{\,s} (\widehat{\mathbf{A}}_i^{\,s})^{\top} - \mathbf{I}_{K} \right\|_F^2,
\end{equation}
where $\widehat{\mathbf{A}}_i^{\,s}$ denotes the row-normalized anchor matrix, with its $k$-th row computed as  $\widehat{\mathbf{a}}_{i,k}^{\,s} = \widetilde{\mathbf{a}}_{i,k}^{\,s} / \|\widetilde{\mathbf{a}}_{i,k}^{\,s}\|_2$, and $\|\cdot\|_F$ represents the Frobenius norm. 

To improve interpretability and mitigate overfitting, we apply an $\ell_1$-norm sparsity constraint. This encourages each anchor to activate only a compact subset of feature dimensions, which can be expressed by
\begin{equation}
    \mathcal{L}_{\mathrm{sparse}}^{s} = \frac{1}{K D} \sum_{k=1}^{K} \|\widetilde{\mathbf{a}}_{i,k}^{\,s}\|_1.
\end{equation}

The total anchor regularization loss is aggregated across all three magnification levels, which can be formulated as follows:
\begin{equation}
    \mathcal{L}_{\mathrm{anchor}} = \sum_{s\in\{5\times,10\times,20\times\}} \big( \mathcal{L}_{\mathrm{orth}}^{s} + \blue{\beta} \mathcal{L}_{\mathrm{sparse}}^{s} \big),
\end{equation}
\blue{where $\beta$ is a hyperparameter that balances the contribution of the sparsity penalty.}


\subsection{Hierarchical Mixture-of-Experts Strategy} 


\blue{Given the aligned cross-modal features $\{U_i^s\}_s$, H-MoE performs two-stage selection and fusion. Specifically, the intra-scale expert selection sparsely routes and filters multimodal representations within each resolution, while the inter-scale MoE fusion dynamically recalibrates and integrates evidence across resolutions. This design adaptively prioritizes the clinically relevant magnifications for each patient}.


\subsubsection{Intra-scale Experts Scoring and Selection}



\blue{We assess each descriptor's prognostic relevance via an intra-scale scoring MoE. Given the complexity of tumor heterogeneity, a single linear evaluator is insufficient; we instead employ $E$ lightweight MLP experts $\{f_e\}_{e=1}^{E}$, each specialized in distinct risk patterns. A sparse top-$k_e$ routing mechanism dynamically assigns each descriptor $\mathbf{u}_{i,k}^{s}$ to the most competent experts}. The final score $r_{i,k}^s$ is obtained via a routed ensemble as follows:
\begin{equation}
    r_{i,k}^s = \sum_{e \in \mathcal{T}_{i,k}^s} \pi_{i,k,e}^s \cdot f_{e}(\mathbf{u}_{i,k}^s),
\end{equation}
where $\mathcal{T}_{i,k}^s$ denotes the set of indices for the $k_e$ experts with the highest routing scores, and $f_e(\cdot)$ maps the descriptor to a scalar confidence score. The gating coefficient $\pi_{i,k,e}^s$, representing the routing probability of the $e$-th expert for the current descriptor, is derived via a routing network as follows:
\begin{equation}
    \pi_{i,k}^s = \mathrm{softmax} \left( \text{Top-}k_e (\mathbf{W}_{\mathrm{gate}}^s \mathbf{u}_{i,k}^s ) \right) \in \mathbb{R}^E,
\end{equation}
where $\mathbf{W}_{\mathrm{gate}}^s \in \mathbb{R}^{E \times D}$ is the routing weight matrix. \blue{The $\mathrm{Top}\mbox{-}k_e(\cdot)$ operation retains the $k_e$ largest routing logits and masks the remaining logits}.


Based on these comprehensive risk scores, we employ a hard selection strategy to isolate the most salient features. We sort the descriptors in descending order of $r_{i,k}^s$ and retain the top-$K'$ ranked candidates:
\begin{equation}
    \mathcal{S}_i^s = \left\{ \mathbf{u}_{i,(1)}^s, \mathbf{u}_{i,(2)}^s, \dots, \mathbf{u}_{i,(K')}^s \right\},
\end{equation}
where the parenthesized subscript $(j)$ denotes the index of the descriptor with the $j$-th highest score. The selected set $S_i^s$ is passed to the inter-scale fusion stage. 

\subsubsection{Inter-Scale Mixture-of-Experts Fusion}

To synthesize the multi-scale evidence, we adopt a Rank-based alignment strategy. We posit that descriptors ranking at the same importance level across different scales share a latent semantic correlation. We construct a multi-scale feature vector $\mathbf{x}_{i,j}$ by concatenating the $j$-th ranked descriptors from all scales:
\begin{equation}
    \mathbf{x}_{i,j} = \left[ \mathbf{u}_{i,(j)}^{5\times} \parallel \mathbf{u}_{i,(j)}^{10\times} \parallel\mathbf{u}_{i,(j)}^{20\times} \right] \in \mathbb{R}^{3D}, \quad j=1,\dots,K'.
\end{equation}


\blue{To mitigate optimization difficulties from direct linear projection of high-dimensional heterogeneous features, we introduce a feature-level MoE for dynamic channel-wise recalibration. It generates a context-aware mask $\mathbf{m}_{i,j}$ for each multi-scale vector $\mathbf{x}_{i,j}$ as a weighted sum of $M$ fusion experts as follows:}
\begin{equation}
    \mathbf{m}_{i,j} = \sum_{m=1}^{M} \eta_{i,j}^{m} \cdot \sigma\left( g_m(\mathbf{x}_{i,j}) \right) \in \mathbb{R}^{3D},
\end{equation}
\blue{where $g_m(\cdot) : \mathbb{R}^{3D} \rightarrow \mathbb{R}^{3D}$ represents the $m$-th mask-generation expert, $\sigma(\cdot)$ is the sigmoid activation, and the routing coefficient vector $\boldsymbol{\eta}_{i,j} = [\eta_{i,j}^1, \ldots, \eta_{i,j}^M]^\top$ is dynamically derived via a standard softmax router: $\boldsymbol{\eta}_{i,j} = \mathrm{softmax}\left( \text{Router}(\mathbf{x}_{i,j}) \right) \in \mathbb{R}^M$. The final fused representation $\mathbf{z}_{i,j}$ is obtained by applying the generated mask to the input features and projecting them back to the latent space:}
\begin{equation}
    \mathbf{z}_{i,j} = \mathbf{W}_{\mathrm{red}} \left( \mathbf{x}_{i,j} \odot \mathbf{m}_{i,j} \right) \in \mathbb{R}^D,
\end{equation}
where $\odot$ denotes element-wise multiplication and $\mathbf{W}_{\mathrm{red}} \in \mathbb{R}^{D \times 3D}$ is a learnable reduction matrix. The patient-level holistic representation is then derived by averaging these fused tokens: $\mathbf{h}_i = \frac{1}{K'} \sum_{j=1}^{K'} \mathbf{z}_{i,j}$.

\subsection{Total Objective Function}

After obtaining the patient-level holistic representation $\mathbf{h}_{i}$ from the H-MoE module, we feed it into the survival prediction head. 
The vector of interval-wise hazard probabilities $\mathbf{p}_{i} = [p_{i,1}, \ldots, p_{i,R}]^\top$ is predicted via a fully connected layer with sigmoid activation:
\begin{equation}
    \label{eq:hazard_probabilities}
   \mathbf{p}_{i} = \sigma(\mathbf{W}_{\text{risk}}\mathbf{h}_{i} + \mathbf{b}),
\end{equation}
where $\mathbf{W}_{\text{risk}} \in \mathbb{R}^{R \times D}$ and $\mathbf{b} \in \mathbb{R}^{R}$ are learnable parameters.

The training objective combines the survival loss with anchor regularization. For patient $i$, with observed time $t_i$ assigned to interval $y_i$ and censoring indicator $c_i$, the survival loss is defined as follows:
\begin{equation}
\begin{aligned}
    \mathcal{L}_{\text{surv}} = -\frac{1}{N} \sum_{i=1}^{N} \bigg[ & (1 - c_i) \log\big(p_{i, y_i} S_{i}(y_i-1)\big) \\
    & + c_i \log\big(S_{i}(y_i)\big) \bigg],
\end{aligned}
\end{equation}
where $S_{i}(\cdot)$ is derived from Eq.~(\ref{eq:survival_function}), and $N$ denotes the number of patients. 

The final objective function is a weighted combination of the survival loss and the anchor regularization terms. Therefore, the total loss function can be expressed by
\begin{equation}
    \mathcal{L}_{\text{total}} = \mathcal{L}_{\text{surv}} + \lambda \mathcal{L}_{\text{anchor}},
\end{equation}
where $\lambda$ is a hyperparameter balancing the prediction task and the regularization constraints.

\section{EXPERIMENTS AND RESULTS}\label{sec:Experiments}

\subsection{Experimental Setup}

\subsubsection{Datasets}

To evaluate the performance of our proposed framework, we conducted extensive experiments on five publicly available cancer datasets from The Cancer Genome Atlas (TCGA), covering both survival outcomes and paired modalities: diagnostic Whole Slide Images (WSIs) and genomic profiles. The datasets include Bladder Urothelial Carcinoma (BLCA), Breast Invasive Carcinoma (BRCA), Uterine Corpus Endometrial Carcinoma (UCEC), Glioblastoma \& Lower Grade Glioma (GBMLGG), and Lung Adenocarcinoma (LUAD). The sample size $N$ is detailed in Table~\ref{tab:model_comparison}. For genomic data preprocessing, we filter genes based on variability and missingness. Following the protocols in~\cite{liberzon2015molecular, MCAT_2021_ICCV, MoCAT_2023_ICCV}, we organize genomic features into $M_g=6$ functional categories based on biological pathways: (1) Tumor Suppression, (2) Oncogenesis, (3) Protein Kinases, (4) Cellular Differentiation, (5) Transcription, and (6) Cytokines and Growth.

\subsubsection{Experimental Settings}

For WSI preprocessing, we employ the OTSU's thresholding algorithm to create tissue masks, effectively filtering out background and artifacts. Non-overlapping patches of size $224 \times 224$ are extracted from the segmented tissue regions at three distinct magnification levels: $5\times$, $10\times$, and $20\times$, to capture hierarchical morphological details. Each patch is encoded using the pretrained UNI model~\cite{UNI_2024} to extract 1024-dimensional embeddings, utilizing the feature extraction pipeline from CLAM~\cite{CLAM_lu2021}, and we set $D=256$ in the latent space. Following our proposed method, we aggregate these dense patch embeddings into scale-specific semantic prototypes using a Transformer-based clustering module. For genomic data, we employ Self-Normalizing Neural Networks (SNNs)~\cite{SNN_NIPS2017} to instantiate the category-specific encoders $f_j(\cdot)$ described in Sec.~\ref{sec:genomic_learning}, following the setting of~\cite{MoCAT_2023_ICCV}.

\begin{table*} 
    \centering
    \renewcommand{\arraystretch}{1.2}
    \setlength\tabcolsep{4.2pt}
    \small
    \caption{Quantitative comparison of C-index (mean $\pm$ std) across five TCGA cohorts. Baselines are categorized by input modalities (g: genomic, h: histology) and WSI scales. The \textbf{best} results are in bold, and the \underline{second-best} are underlined. 
	}\label{tab:model_comparison}
    \begin{tabular}{r|ll|ccccc|c}
    \toprule
    \multirow{2}{*}{Methods} & \multirow{2}{*}{Modal} & \multirow{2}{*}{Scale} & BLCA & BRCA & GBMLGG & LUAD & UCEC & \multirow{2}{*}{Average} \\ 
    & & & (N=373) & (N=955) & (N=550) & (N=452) & (N=480) & \\ 
    \midrule
    MLP~      \cite{MLP_1994neural}      & g.   & - 
    & 0.613 $\pm$ 0.019   & 0.587 $\pm$ 0.033    & 0.809 $\pm$ 0.029 
    & 0.617 $\pm$ 0.026   & 0.657 $\pm$ 0.036    & 0.657 \\
    SNN~      \cite{SNN_NIPS2017}        & g.   & - 
    & 0.619 $\pm$ 0.023   & 0.596 $\pm$ 0.027    & 0.805 $\pm$ 0.030
    & 0.625 $\pm$ 0.019   & 0.651 $\pm$ 0.018    & 0.659 \\
    SNNTrans~ \cite{SNNTrans_2023_ICCV}  & g.   & - 
    & 0.627 $\pm$ 0.019   & 0.618 $\pm$ 0.018    & 0.816 $\pm$ 0.037
    & 0.631 $\pm$ 0.023   & 0.641 $\pm$ 0.026    & 0.667 \\
    ABMIL~    \cite{ABMIL_pmlr}          & h.   & 20$\times$  
    & 0.622 $\pm$ 0.051   & 0.614 $\pm$ 0.037    & 0.786 $\pm$ 0.028
    & 0.596 $\pm$ 0.067   & 0.649 $\pm$ 0.028    & 0.652 \\
    CLAM~     \cite{CLAM_lu2021}         & h.   & 20$\times$  
    & 0.613 $\pm$ 0.058   & 0.607 $\pm$ 0.016    & 0.792 $\pm$ 0.032
    & 0.590 $\pm$ 0.076   & 0.650 $\pm$ 0.066    & 0.650 \\
    TransMIL~ \cite{TransMIL_2021}       & h.   & 20$\times$
    & 0.625 $\pm$ 0.050   & 0.614 $\pm$ 0.037    & 0.803 $\pm$ 0.037
    & 0.593 $\pm$ 0.063   & 0.665 $\pm$ 0.019    & 0.660 \\
    WiKG~     \cite{WiKG_2024_CVPR}      & h.   & 20$\times$ 
    & - & 0.507 $\pm$ 0.001 & - & 0.592 $\pm$ 0.008 & 0.562 $\pm$ 0.001 & 0.553 \\
    \midrule
    DSMIL~    \cite{DSMIL_2021_CVPR}     & h.   & 5$\times$20$\times$ 
    & - & 0.547 $\pm$ 0.055 & - & 0.588 $\pm$ 0.083 & 0.613 $\pm$ 0.039 & 0.582 \\
    CSMIL~    \cite{CSMIL_2024_MIA}      & h.   & 5$\times$10$\times$20$\times$ 
    & - & 0.541 $\pm$ 0.029 & - & 0.550 $\pm$ 0.069 & 0.526 $\pm$ 0.082 & 0.539 \\
    HIPT~     \cite{HIPT_2022_CVPR}      & h.   & 5$\times$10$\times$20$\times$ 
    & - & 0.538 $\pm$ 0.087 & - & 0.558 $\pm$ 0.031 & 0.602 $\pm$ 0.042 & 0.566 \\
    \midrule
    PORPOISE~ \cite{PORPOISE2022pan}     & h.$+$g.   & 20$\times$ 
    & 0.661 $\pm$ 0.034   & 0.644 $\pm$ 0.010   & 0.829 $\pm$ 0.033 
    & 0.646 $\pm$ 0.025   & 0.660 $\pm$ 0.106   & 0.688 \\
    MCAT~     \cite{MCAT_2021_ICCV}      & h.$+$g.   & 20$\times$ 
    & 0.660 $\pm$ 0.046   & 0.652 $\pm$ 0.019   & 0.833 $\pm$ 0.029
    & 0.642 $\pm$ 0.054   & 0.686 $\pm$ 0.042   & 0.694 \\
    MoCAT~     \cite{MoCAT_2023_ICCV}    & h.$+$g.   & 20$\times$ 
    & \underline{0.667 $\pm$ 0.038}   & 0.659 $\pm$ 0.034   & 0.835 $\pm$ 0.037 
    & 0.656 $\pm$ 0.021   & 0.683 $\pm$ 0.060   & 0.700 \\
    Survpath~  \cite{Survpath_2024_CVPR} & h.$+$g.   & 20$\times$ 
    & 0.663 $\pm$ 0.026   & \underline{0.661 $\pm$ 0.024} & \underline{0.839 $\pm$ 0.039}
    & \underline{0.671 $\pm$ 0.036}   & \underline{0.709 $\pm$ 0.039} & \underline{0.709} \\
    
    \midrule
    M3IF~      \cite{M3IF_2021multi}     & h.$+$g.   & 5$\times$10$\times$20$\times$ 
    & - & 0.528 $\pm$ 0.060 & - & 0.614 $\pm$ 0.045 & 0.615 $\pm$ 0.060  & 0.585 \\
    HiMT~      \cite{HiMT_2022}          & h.$+$g.   & 5$\times$10$\times$20$\times$ 
    & 0.660 $\pm$ 0.021 & 0.490 $\pm$ 0.066 & 0.823 $\pm$ 0.019 & 0.577 $\pm$ 0.018 & 0.583 $\pm$ 0.089 & 0.626 \\
    UMSA~      \cite{UMSA_2025}          & h.$+$g.   & 5$\times$10$\times$20$\times$ 
    & - & 0.576 $\pm$ 0.050 & - & 0.652 $\pm$ 0.083 & 0.649 $\pm$ 0.046 & 0.625 \\
    Ours                                 & h.$+$g.   & 5$\times$10$\times$20$\times$ 
    & \textbf{0.685 $\pm$ 0.016}   & \textbf{0.693 $\pm$ 0.040}   & \textbf{0.849 $\pm$ 0.019} 
    & \textbf{0.687 $\pm$ 0.020}   & \textbf{0.762 $\pm$ 0.028}   & \textbf{0.735} \\
    \bottomrule

\end{tabular}
\end{table*} 

\begin{figure*}[!t]
	\centering
    \includegraphics[width=0.95\textwidth]{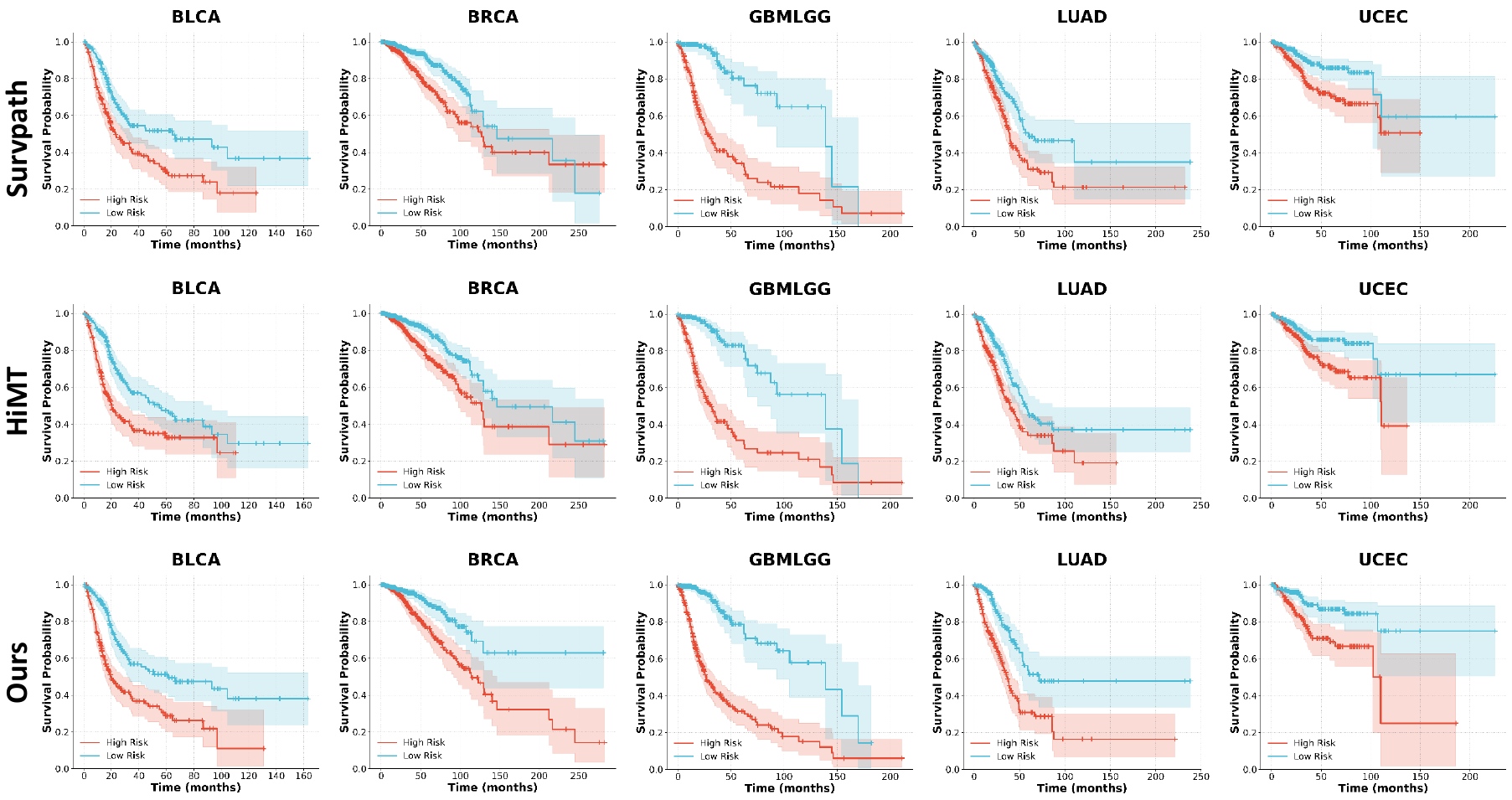} 
    \vspace{-0.2cm} 
	\caption{\footnotesize Kaplan-Meier survival analysis across five cancer cohorts (BLCA, BRCA, GBMLGG, LUAD, and UCEC), comparing our proposed AM$^2$ES framework against Survpath and HiMT. Patients are stratified into high-risk (red) and low-risk (blue) groups based on the median predicted risk score. The shaded regions represent $95\%$ confidence intervals.}
    \label{fig:km_curves}
\end{figure*}

\subsubsection{Implementation Details and Evaluation Metric}

All experiments are conducted in a PyTorch 2.6.0 environment with CUDA 12.4 and an NVIDIA GeForce RTX 4090 GPU with 24GB memory. During model training, we adopt the Adam optimizer with an initial learning rate of $2 \times 10^{-4}$ and a weight decay of $1 \times 10^{-5}$. \blue{All experiments are trained for 20 epochs. For the WSI representation learning, the number of scale-specific semantic prototypes $P^s$ is set to 64, 128, and 256 for the $5\times$, $10\times$, and $20\times$ magnifications, respectively. For the proposed AM$^2$ES architecture, the number of scale-specific dynamic anchors is set to $K=16$. In the H-MoE module, we configure $E=4$ intra-scale scoring experts with a sparse top-$k_e=2$ gating strategy, and retain the top $K'=4$ candidates per scale. This is followed by $M=4$ inter-scale experts for feature recalibration. Additionally, we empirically set $\lambda=0.05$ and ${\beta}=0.1$}. 
For a fair evaluation, we standardized the configurations of all comparison methods in our experiments. 

Moreover, we employ a 5-fold cross-validation strategy with patient-level separation. This ensures that all WSI patches belonging to the same patient are exclusively assigned to either the training or testing set, strictly preventing data leakage. To evaluate survival prediction performance, we employ the Concordance Index (C‑index), a standard metric that quantifies the model's ability to correctly rank patient risk scores relative to their observed survival outcomes.


\subsection{Comparison with State-of-the-art Methods}

\subsubsection{Comparison Methods}

To demonstrate the effectiveness and robustness of our proposed framework, we conduct a comprehensive comparison against a wide range of state-of-the-art (SOTA) methods. As reported in Table~\ref{tab:model_comparison}, these baselines are rigorously categorized based on input modality (Unimodal vs. Multimodal) and visual granularity (Single-scale vs. Multi-scale). 
\textbf{Unimodal Methods}: For genomic data only, we utilize MLP~\cite{MLP_1994neural}, SNN~\cite{SNN_NIPS2017}, and transformer-based SNNTrans~\cite{SNNTrans_2023_ICCV} as benchmarks. For WSI data only, we compare the proposed model with methods across different scales. For single-scale settings ($20\times$), we include standard MIL approaches: ABMIL~\cite{ABMIL_pmlr}, CLAM~\cite{CLAM_lu2021}, TransMIL~\cite{TransMIL_2021}, and WiKG~\cite{WiKG_2024_CVPR}. To assess multi-scale capabilities ($5\times, 10\times, 20\times$), we compare against DSMIL~\cite{DSMIL_2021_CVPR}, CSMIL~\cite{CSMIL_2024_MIA}, and the hierarchical transformer HIPT~\cite{HIPT_2022_CVPR}. 
\textbf{Multimodal Methods}: We further compare our model against SOTA multimodal fusion frameworks. 
The first group operates on single-scale WSIs, including PORPOISE~\cite{PORPOISE2022pan}, MCAT~\cite{MCAT_2021_ICCV}, MoCAT~\cite{MoCAT_2023_ICCV}, and Survpath~\cite{Survpath_2024_CVPR}. 
The second group involves multi-scale fusion methods, specifically M3IF~\cite{M3IF_2021multi}, HiMT~\cite{HiMT_2022}, and UMSA~\cite{UMSA_2025}, which serve as the most direct competitors to our hierarchical approach.
For a fair comparison, all WSI-based methods utilize the same UNI-encoded features and identical data splits.

\subsubsection{Result Comparison}

The quantitative results of survival prediction across five TCGA cohorts are summarized in Table~\ref{tab:model_comparison}. As observed, our model achieves the best overall performance, demonstrating superior generalization and robustness compared to existing SOTA methods.

\textbf{Comparison with Unimodal Methods}: 
In Table~\ref{tab:model_comparison}, we compare the proposed model against gene-based baselines (MLP, SNN, SNNTrans) and several pathology-based approaches.
Among genomic methods, the transformer-based SNNTrans achieves the highest C-index of $0.667$, outperforming MLP ($0.657$) and SNN ($0.659$). This suggests that attention mechanisms are effective at capturing complex intra-genomic interactions. In the pathological domain, the transformer-based TransMIL leads with a C-index of $0.660$, surpassing ABMIL ($0.652$) and CLAM ($0.650$). Although multi-scale unimodal methods such as HIPT and DSMIL aim to incorporate hierarchical information, their performance (\eg, HIPT on BRCA: $0.538$) does not consistently exceed that of single-scale transformer-based approaches. This suboptimal result may stem from the lack of effective cross-modal guidance to filter out irrelevant features across different magnifications.

\begin{table}[!t]
\centering
\small
 \renewcommand{\arraystretch}{1.1}
 \setlength\tabcolsep{6.2pt}
\caption{\blue{Comparison of external validation performance (C-index) on two independent CPTAC cohorts}. 
}

\label{tab_external}
\begin{tabular}{r|cc}
\toprule
\textbf{Method} & \textbf{CPTAC-UCEC\cite{CPTAC_UCEC_2019}} & \textbf{CPTAC-LUAD}~\cite{CPTAC_LUAD_2018} \\
\midrule
Survpath~\cite{Survpath_2024_CVPR}  & 0.5034 $\pm$ 0.0118   & 0.5670 $\pm$ 0.0341 \\
HiMT~\cite{HiMT_2022}               & 0.5470 $\pm$ 0.0747   & 0.5269 $\pm$ 0.0538 \\
\textbf{Ours}         & \textbf{0.5651 $\pm$ 0.0690} & \textbf{0.5843 $\pm$ 0.0428} \\
\bottomrule
\end{tabular}
\end{table}

\textbf{Comparison with Multimodal Methods}: 
Compared to unimodal approaches, multimodal methods generally demonstrate superior performance in survival prediction. Methods that emphasize modality alignment, such as MoCAT and MCAT, achieve C-index scores of $0.700$ and $0.694$, respectively. Survpath further improves upon this by leveraging topological interactions, reaching a C-index of 0.709. However, these approaches are limited by their reliance on single-scale inputs ($20\times$), which fail to capture broader contextual tissue structures. Among multi-scale fusion methods, HiMT and UMSA exhibit competitive performance but achieve only average C-indices of $0.626$ and $0.625$, respectively, likely due to their relatively simplistic fusion or alignment strategies. Notably, our method attains a superior average C-index of $0.735$, substantially outperforming the second-best method, Survpath ($0.709$). Specifically on the UCEC dataset, our AMF-MES model achieves an impressive score of $0.762$, representing a considerable margin over Survpath’s $0.709$. This consistent performance gain underscores the critical role of our anchor-guided routing mechanism, which adaptively integrates biological pathway information with hierarchical visual cues.

\begin{table*}[!t]
\centering
\caption{\textbf{Comprehensive ablation study} of AM$^2$ES. The study validates: (a) input modalities, (b) multi-scale hierarchy, (c) anchor fusion mechanism, (d) hierarchical expert mechanism, and \blue{(e) different inter-scale alignment strategies}. 
}
\label{tab:ablation}

\renewcommand{\arraystretch}{1.2} 
\setlength{\tabcolsep}{12pt}

\resizebox{0.9\textwidth}{!}{%
\begin{tabular}{c|l|ccc|c} 
    \hline
    \textbf{ID} & \textbf{Model Variants} & \textbf{BRCA} & \textbf{LUAD} & \textbf{UCEC} & \textbf{Avg.} \\ 
    \hline \hline 

    \multicolumn{6}{l}{\textit{\textbf{(a) Impact of Input Modalities}}} \\
    1 & g.~(MLP)    & 0.587 $\pm$ 0.033  & 0.617 $\pm$ 0.026  & 0.657 $\pm$ 0.036 & 0.620 \\
    2 & h.~(ABMIL)  & 0.614 $\pm$ 0.037  & 0.596 $\pm$ 0.067  & 0.649 $\pm$ 0.028 & 0.620 \\
    3 & h.+g. (Concat) & 0.630 $\pm$ 0.021 & 0.632 $\pm$ 0.047 & 0.689 $\pm$ 0.032 & 0.650 \\ 
    \hline

    \multicolumn{6}{l}{\textit{\textbf{(b) Impact of Multi-scale Hierarchy}}} \\
    4 & Single Scale ($20\times$) & 0.663 $\pm$ 0.024 & 0.644 $\pm$ 0.039 & 0.743 $\pm$ 0.036 & 0.683 \\
    5 & Dual Scale ($5\times20\times$) & 0.661 $\pm$ 0.037 & 0.648 $\pm$ 0.031  & 0.752 $\pm$ 0.042 & 0.687 \\
    6 & Full Hierarchy ($5\times10\times20\times$) & \textbf{0.693 $\pm$ 0.040} & \textbf{0.687 $\pm$ 0.020}
    & \textbf{0.762 $\pm$ 0.028} & \textbf{0.714} \\ 
    \hline

    \multicolumn{6}{l}{\textit{\textbf{(c) Impact of Anchor Fusion Mechanism}}} \\
    7 & w/ Cross-Attention (TransMIL-style) & 0.663 $\pm$ 0.013 & 0.657 $\pm$ 0.035 
    & 0.746 $\pm$ 0.033 & 0.689 \\
    8 & w/ Bilinear Pooling & 0.650 $\pm$ 0.035 & 0.650 $\pm$ 0.036 & 0.745 $\pm$ 0.030 & 0.682 \\ 
    \hline

    \multicolumn{6}{l}{\textit{\textbf{(d) Impact of Hierarchical Expert Mechanism}}} \\
    \multicolumn{6}{l}{\quad \textit{Stage 1: Intra-scale Selection Strategy}} \\
    9 & w/ Global Average Pooling (No Selection) & 0.654 $\pm$ 0.031 & 0.653 $\pm$ 0.032 & 0.751 $\pm$ 0.025 & 0.686 \\
    10 & w/ Attention Pooling (Soft Selection) & 0.667 $\pm$ 0.054 & 0.657 $\pm$ 0.030 & 0.748 $\pm$ 0.020 & 0.691 \\
    \hline
    \multicolumn{6}{l}{\quad \textit{Stage 2: Inter-scale Fusion Strategy}} \\
    11 & w/ Linear Concatenation (No MoE) & 0.673 $\pm$ 0.026 & 0.656 $\pm$ 0.018 & 0.733 $\pm$ 0.51 & 0.687 \\
    12 & \textbf{AMF-MES (Top-K Selection + MoE Routing)} & \textbf{0.693 $\pm$ 0.040} & \textbf{0.687 $\pm$ 0.020} & \textbf{0.762 $\pm$ 0.028} & \textbf{0.714} \\ 
    \hline

    \multicolumn{6}{l}{\textit{\textbf{(e) \blue{Impact of Different Inter-scale Alignment Strategies }}}} \\
    13 & No Alignment (Global Pooling)    & 0.651 $\pm$ 0.032  & 0.680 $\pm$ 0.028   & 0.734 $\pm$ 0.029   & 0.688 \\
    14 &Original Order (w/o Sorting)  & 0.657 $\pm$ 0.035  & 0.673 $\pm$ 0.029   & 0.750 $\pm$ 0.031   & 0.693 \\
    15 & Random Alignment            & 0.670 $\pm$ 0.033  & 0.652 $\pm$ 0.028   & 0.741 $\pm$ 0.030   & 0.688 \\ 
    16 & \textbf{Rank-based Alignment (Ours)}  & \textbf{0.693 $\pm$ 0.040}    
    & \textbf{0.687 $\pm$ 0.020}   & \textbf{0.762 $\pm$ 0.028}   & \textbf{0.714} \\ 
    \hline

\end{tabular}%
}
\vspace{-0.05cm}
\end{table*}

\blue{\textbf{External Validation}. 
We further evaluate our model on two independent CPTAC cohorts: Uterine Corpus Endometrial Carcinoma (CPTAC-UCEC~\cite{CPTAC_UCEC_2019}) and Lung Adenocarcinoma (CPTAC-LUAD~\cite{CPTAC_LUAD_2018}). To ensure a fair cross-cohort evaluation, we apply data harmonization and quality-control procedures consistent with the TCGA training setting. Only patients with complete multi-omics profiles and valid survival data were included; cases with zero survival time were excluded. The final cohorts comprised 34 CPTAC-UCEC and 53 CPTAC-LUAD patients. For external validation, models trained on the corresponding TCGA cohorts are directly applied to the CPTAC cohorts without fine-tuning or retraining. We compare our method with two representative approaches, \ie, SurvPath~\cite{Survpath_2024_CVPR} and HiMT~\cite{HiMT_2022}. As shown in Table \ref{tab_external}, our model achieves the best performance on both cohorts. Although absolute C-index values were lower than in the internal TCGA evaluations, this is expected in zero-shot cross-cohort validation owing to batch effects, cohort differences, and limited sample sizes. Nevertheless, our method consistently outperformed competing approaches on both CPTAC cohorts, supporting its robustness and generalizability. }



\subsubsection{Kaplan-Meier Survival Analysis}

To further assess the clinical value of our model in risk stratification, we perform Kaplan-Meier (KM) survival analysis on five TCGA cancer cohorts. Patients in each cohort are stratified into high-risk and low-risk groups based on the median predicted risk score derived from 5-fold cross-validation.


As illustrated in Fig.~\ref{fig:km_curves}, the Kaplan-Meier curves reveal statistically significant risk stratification across all cancer types. Visually, the high-risk (red) and low-risk (blue) curves diverge clearly shortly after the initial time point. In aggressive cancer subtypes such as BLCA and GBMLGG, the high-risk group exhibits a steep decline in survival probability within the first 20–40 months, whereas the low-risk group maintains a relatively stable survival rate. Furthermore, Fig.~\ref{fig:km_curves} provides a comparison between our model and two strong baselines, \ie, Survpath and HiMT. As the baseline methods struggle to maintain clear separation in cohorts like BRCA and UCEC, frequently exhibiting overlapping confidence intervals (shaded regions) in the later stages, while our framework achieves a consistently wider and more stable margin between the risk groups. This clear separation, accompanied by minimal overlap in the $95\%$ confidence intervals, further underscores the model's robustness. These results confirm that our model can effectively disentangle risk profiles at an early stage, highlighting its superior clinical utility in guiding timely therapeutic interventions compared to existing methods.

\subsection{Ablation Study}

To systematically evaluate the contribution of each component in our proposed framework and verify the robustness of our design choices, we conduct extensive ablation experiments on the BRCA, LUAD, and UCEC datasets. The ablative results are summarized in Table~\ref{tab:ablation}. 

\subsubsection{Effectiveness of Each Component} 

\textbf{(a) Impact of Input Modalities.}
We first investigate the necessity of multimodal integration. As shown in Table~\ref{tab:ablation} (IDs 1-3), the uni-modal baselines yield suboptimal performance, with both the genomic-only (MLP) and WSI-only (ABMIL) models achieving an identical average C-index of $0.620$. Simply concatenating features from both modalities (ID 3) brings a noticeable improvement ($+3.0\%$), reaching $0.650$. This confirms that genomic profiles and pathological images contain complementary prognostic signals. However, the performance of simple concatenation is still limited compared to our full model ($0.714$), highlighting the importance of our advanced fusion module.

\textbf{(b) Impact of Multi-scale Hierarchy.}
We further assess the contribution of the hierarchical processing of WSIs. In Exp. (4)-(6) in Table~\ref{tab:ablation}, we evaluate different scale combinations. The single-scale model ($20\times$) achieves an average C-index of $0.683$. Incorporating the $5\times$ scale (Dual Scale, ID 5) slightly improves the performance to $0.687$. The best performance is achieved by utilizing the full hierarchy ($5\times, 10\times, 20\times$) in Exp. (6), reaching a C-index of $0.708$. This result validates our hypothesis that a holistic survival analysis requires multi-granular visual information, combining global tissue architecture with local cellular details.

\textbf{(c) Impact of Multi-modal Fusion Strategy.} A key contribution of the proposed framework is the dynamic anchor-driven fusion strategy. To verify its superiority, we compare it against two established fusion strategies under the same multi-scale setting: Cross-Attention (widely used in TransMIL-style models) and Bilinear Pooling. As reported in experiments (7) and (8) in Table~\ref{tab:ablation}, our method ($0.714$) outperforms Cross-Attention ($0.689$) and Bilinear Pooling ($0.682$) by margins of $2.5\%$ and $3.2\%$, respectively. We attribute this improvement to the design of patient-specific anchors, which act as dynamic prototypes that align high-dimensional WSI features with genomic embeddings more effectively than generic attention maps, thereby reducing semantic misalignment.


\textbf{(d) Impact of MoE Routing.}
Finally, we analyze the effectiveness of our hierarchical expert design, dissecting it into two stages: intra-scale selection and inter-scale fusion. \textit{\textbf{Stage 1: Intra-scale Selection.}} We compare the Top-K instance selection strategy against Global Average Pooling (GAP) and Attention Pooling. As shown in entries (9)-(10) in Table~\ref{tab:ablation}, GAP (No Selection) and Attention Pooling (Soft Selection) yield average C-indices of 0$.686$ and $0.691$, respectively. Our Top-K strategy (embedded in ID 12) achieves the highest performance ($0.714$), demonstrating that filtering out non-informative patches explicitly is more effective for prognosis than soft weighting or global averaging. \textit{\textbf{Stage 2: Inter-scale Fusion.}} We further verify the necessity of the MoE module for fusing features across scales. Replacing the MoE dynamic gating with a static Linear Concatenation (shown in entry (11)) results in a performance drop to $0.687$. This demonstrates that the ``soft" routing mechanism in MoE allows the model to adaptively weight contributions from different magnifications based on sample difficulty, rather than relying on a fixed combination.

\begin{figure*}[!t]
	\centering
	\footnotesize
    \includegraphics[width=0.999\textwidth]{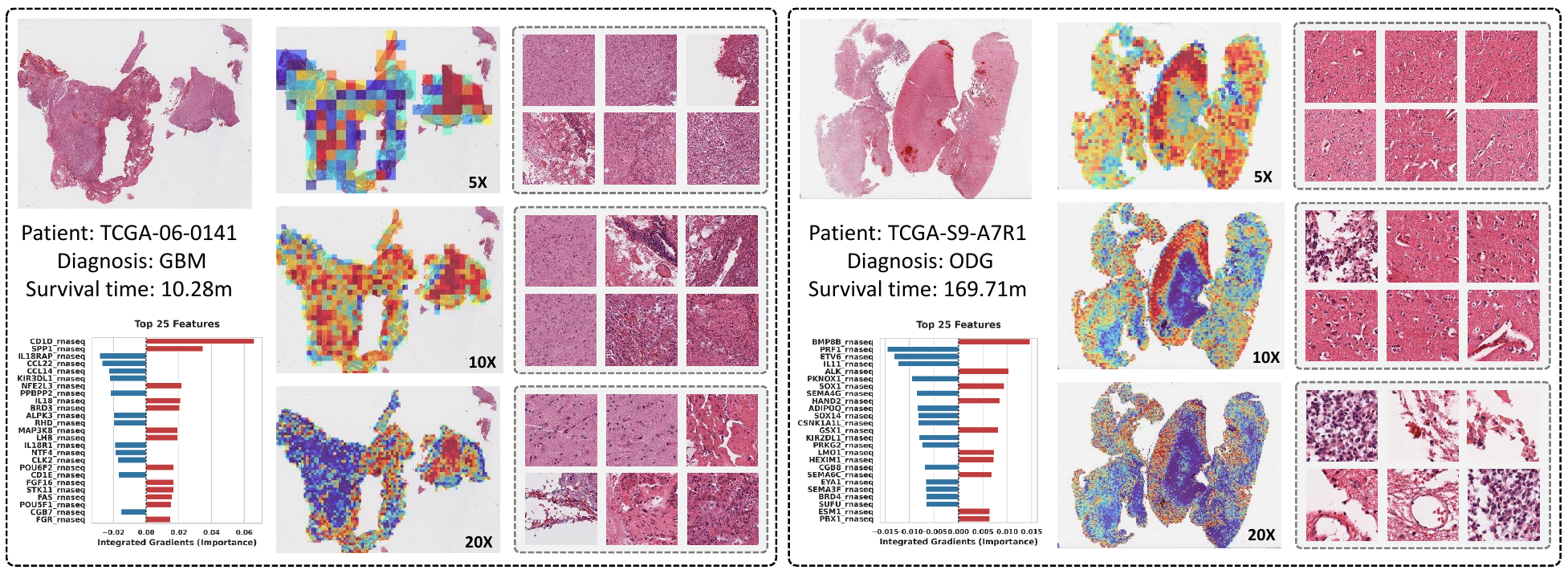} 
    \vspace{-0.2cm}
	\caption{\footnotesize 
        Visualization of multimodal interpretability for glioma survival prediction. This compares a high-risk GBM patient (Left, TCGA-06-0141, survival 10.28 months) and a low-risk ODG patient (Right, TCGA-S9-A7R1, survival 169.71 months). 
        For each patient, the visualization comprises: 
        (1) The Whole Slide Image (WSI) overlaid with spatial attention heatmaps, highlighting regions contributing most to the prediction; 
        (2) Representative pathological patches extracted from these high-attention regions across three distinct magnification scales (as labeled: $5\times$, $10\times$, and $20\times$), capturing both macroscopic tumor architecture and microscopic cellular details; 
        and (3) The top 25 genomic features ranked by importance using Integrated Gradients, where red and blue bars indicate positive and negative contributions to the risk prediction, respectively.
    }
	\label{fig5}
\end{figure*}

\begin{table}[!t]
    \centering
    \renewcommand{\arraystretch}{1.1} 
    \setlength{\tabcolsep}{6pt} 
    \caption{Sensitivity analysis of anchor number $K$ across LUAD, UCEC, and BRCA datasets.}
    \label{tab:param_k}  
    \small
    \begin{tabular}{c ccc c}
        \toprule
        \multirow{2}{*}{$K$} & \multicolumn{3}{c}{Dataset (C-index)} & \multirow{2}{*}{Avg.} \\ 
        \cmidrule(lr){2-4} 
          & BRCA & LUAD & UCEC &  \\
        \midrule
        6   & 0.647 $\pm$ 0.023 & 0.660 $\pm$ 0.026 & 0.734 $\pm$ 0.043 & 0.680 \\
        8   & 0.669 $\pm$ 0.014 & 0.660 $\pm$ 0.020 & 0.740 $\pm$ 0.016 & 0.690 \\
        \textbf{16} & \bf{0.693 $\pm$ 0.040}  & \bf{0.687 $\pm$ 0.020}  & \bf{0.762 $\pm$ 0.028}  & \textbf{0.714} \\
        32  & 0.674 $\pm$ 0.014 & 0.678 $\pm$ 0.029 & 0.748 $\pm$ 0.036 & 0.700 \\
        \bottomrule
    \end{tabular}
\end{table}

\blue{\textbf{(e) Impact of Different Inter-scale Alignment Strategies.} To validate our inter-scale alignment strategy, we compare it with three alternatives: 1) \textbf{No Alignment (Global Pooling):} Each scale is independently compressed using global average pooling, thereby discarding fine-grained correspondence among cross-scale descriptors. 2) \textbf{Original Order (w/o Sorting):} The top-$K'$ highest-scoring descriptors are retained but preserve their original index order, \ie, index $j$ corresponds to its original unsorted identifier $k$. This setting directly isolates the contribution of the proposed descending risk-based sorting mechanism. 3) \textbf{Random Alignment:} Descriptor indices at one scale are randomly shuffled before fusion, intentionally disrupting the rank-based risk hierarchy while preserving the feature distribution. As shown in Table~\ref{tab:ablation}, both \textit{Original Order} and \textit{Random Alignment} consistently reduce performance across cohorts, supporting meaningful semantic correspondence among same-ranked descriptors across scales. Our rank-based alignment captures these cross-scale correlations without additional learnable parameters or computational overhead.}


\begin{table}[h]
    \centering
    \renewcommand{\arraystretch}{1.1} 
    \setlength{\tabcolsep}{5pt} 
    \caption{Sensitivity analysis of the retained top instances number $K'$ across BRCA, LUAD, and UCEC datasets.}
    \label{tab:param_k_prime}  
    \small
    \begin{tabular}{c ccc c}
        \toprule
        \multirow{2}{*}{$K'$} & \multicolumn{3}{c}{Dataset (C-index)} & \multirow{2}{*}{Avg.} \\ 
        \cmidrule(lr){2-4} 
          & BRCA & LUAD & UCEC &  \\
        \midrule
        2   & 0.659 $\pm$ 0.036 & 0.653 $\pm$ 0.027 & 0.756 $\pm$ 0.012 & 0.689 \\
        \textbf{4} & \bf{0.693 $\pm$ 0.040}  & \bf{0.687 $\pm$ 0.020} & \bf{0.762 $\pm$ 0.028}  & \textbf{0.714} \\
        8   & 0.669 $\pm$ 0.020 & 0.653 $\pm$ 0.031 & 0.747 $\pm$ 0.024 & 0.690 \\
        16  & 0.671 $\pm$ 0.018 & 0.666 $\pm$ 0.031 & 0.744 $\pm$ 0.023 & 0.694 \\
        \bottomrule
    \end{tabular}
\end{table}

\begin{table}[h]
    \centering
    \renewcommand{\arraystretch}{1.1} 
    \setlength{\tabcolsep}{5pt} 
    \caption{Ablation on the anchor regularization loss ($\mathcal{L}_{\mathrm{anchor}}$).}
    \label{tab:ablation_loss}  
    \footnotesize 
    \begin{tabular}{@{} l cccc @{}} 
        \toprule
        \multirow{2}{*}{Setting} & \multicolumn{3}{c}{Dataset} & \multirow{2}{*}{Avg.} \\ 
        \cmidrule(l){2-4} 
         & BRCA & LUAD & UCEC &  \\
        \midrule
        w/o $\mathcal{L}_{\mathrm{anchor}}$ & 0.678 $\pm$ 0.020 & 0.675 $\pm$ 0.028 & 0.746 $\pm$ 0.033 & 0.700 \\
        \textbf{w/ $\mathcal{L}_{\mathrm{anchor}}$} & \textbf{0.693 $\pm$ 0.040} & \textbf{0.687 $\pm$ 0.020} & \textbf{0.762 $\pm$ 0.028} & \textbf{0.714} \\
        \bottomrule
    \end{tabular}
\end{table}

\subsubsection{Hyperparameter Sensitivity}

To determine the optimal settings for our model, we investigate the effects of key hyperparameters within our model and structural constraints: the number of dynamic anchors ($K$) in the prototype bridge, the number of retained instances per scale $K'$, and the necessity of anchor regularization.

\textbf{Impact of Anchor Quantity ($K$).} 
The anchors act as a semantic bottleneck to align high-dimensional pixel data with gene expression. We tested $K \in \{6, 8, 16, 32\}$. Results in Table~\ref{tab:param_k} indicate that performance improves significantly from $K=6$ to $K=16$. As observed, a smaller $K$ fails to capture the full spectrum of histological subtypes (\eg, endometrioid vs. serous). However, increasing $K$ to $32$ yields diminishing returns. Therefore, $K=16$ provides a sufficient semantic basis for the cross-modal attention bridge while preventing overfitting to patient-specific noise.

\textbf{Impact of Retained Instances ($K'$).} 
This parameter controls the number of most informative patches (instances) selected from each visual scale during the intra-scale selection stage. These selected instances are subsequently fed into the cross-modal fusion module. We evaluate the model with $K' \in \{2, 4, 8, 16\}$ to determine the optimal information bottleneck. As presented in Table~\ref{tab:param_k_prime}, the performance initially improves and peaks at \textbf{$K'=4$}. 
When $K'=2$, the selection is too restrictive, potentially discarding valuable morphological context (e.g., tumor margins, immune infiltrates) that is crucial for accurate prognosis. At \textbf{$K'=4$}, the model achieves the optimal trade-off, successfully capturing the most distinctive pathological patterns without overwhelming the fusion router. However, as $K'$ increases to $8$ or $16$, we observe a slight degradation in performance. This indicates that retaining an excessive number of instances inevitably introduces noisy or uninformative patches (such as background, healthy tissue, or necrosis), which dilutes the attention mechanism and misguides the subsequent genomic alignment. Consequently, we set $K'=4$ as the default setting to ensure a pure and highly discriminative visual representation.

\textbf{Impact of Anchor Regularization.} 
To prevent the learnable anchors from suffering mode collapse or overfitting to noise, we introduce a composite regularization objective $\mathcal{L}_{\mathrm{anchor}}$ (detailed in Sec.~\ref{sec:anchor_reg}). This objective enforces both orthogonality and sparsity to ensure that the anchors capture diverse and compact biological concepts. To validate the overall necessity of this structural constraint, we evaluate the model by completely removing the $\mathcal{L}_{\mathrm{anchor}}$ during training. As shown in Table~\ref{tab:ablation_loss}, omitting the regularization leads to a noticeable performance drop, with the average C-index decreasing from $0.714$ to $0.700$. This confirms that explicitly constraining the anchor space is crucial for learning distinctive representations, ultimately preventing semantic collapse during multi-modal fusion.

\begin{figure*}[!t]
	\centering
	\footnotesize
    \includegraphics[width=\textwidth]{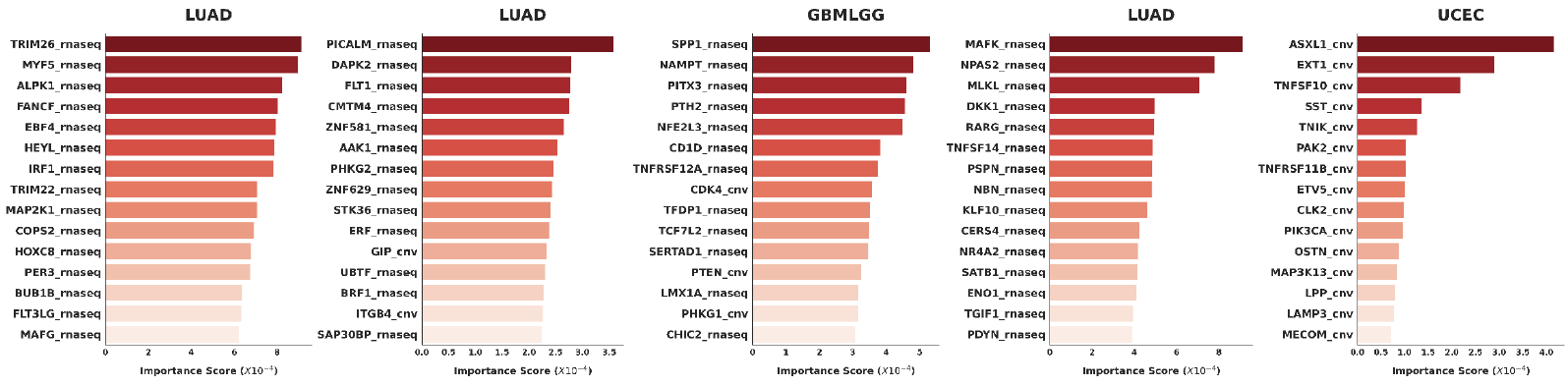}
    \vspace{-0.1cm}
	\caption{\footnotesize 
        Global feature importance analysis across five diverse TCGA pan-cancer datasets. The horizontal bar charts display the top-ranked genomic features (comprising both RNA-seq expression and Copy Number Variations) for each respective cohort, quantified by their overall importance to the model's survival predictions. The identification of completely distinct, disease-specific biomarker signatures across different cohorts demonstrates the model's robust capacity to capture highly generalized, biologically relevant molecular pathways across varied oncology domains.
    }
	\label{fig:inter}\vspace{-0.15cm}
\end{figure*}

\subsection{Interpretability and Case Studies}
\label{sec:interpretability}


To better understand the decision-making process of the proposed multimodal framework, we conduct interpretability analyses for both imaging and genomic modalities. Specifically, spatial attention weights are visualized on whole-slide images (WSIs) to highlight the regions most influential to survival prediction, while feature importance for genomic inputs is quantified using Integrated Gradients (IG). Fig. \ref{fig5} presents a comparative case study of two glioma patients with markedly different clinical outcomes: a high-risk glioblastoma (GBM) patient and a low-risk oligodendroglioma (ODG) patient.

\subsubsection{Morphological Attention Analysis}

As shown in Fig.~\ref{fig:inter}, the spatial attention maps reveal distinct morphological patterns across multiple magnification levels that are consistent with known histopathological characteristics of different glioma subtypes. 
For the GBM patient (TCGA-06-0141, left panel), the attention heatmaps are primarily concentrated within tumor-dense regions of the WSI. Representative patches extracted from these regions illustrate disrupted tissue architecture and increased cellular density at lower magnifications ($5\times$ and $10\times$). At higher magnification ($20\times$), the highlighted regions exhibit morphological features suggestive of high-grade glioma pathology, including pronounced nuclear atypia and heterogeneous cellular morphology. These observations are consistent with the aggressive histological phenotype commonly observed in glioblastoma. 
In contrast, for the ODG patient (TCGA-S9-A7R1, right panel), the attention maps display a more distributed spatial pattern across tumor regions. Across magnification scales, the selected patches consistently exhibit morphological characteristics typical of oligodendroglial tumors. At higher magnification, the model focuses on regions containing relatively uniform round nuclei with subtle perinuclear halos, corresponding to the well-known ``fried-egg” appearance associated with oligodendroglioma. Such morphological patterns are generally associated with more favorable clinical outcomes compared with high-grade gliomas.

Overall, these multi-scale visualizations suggest that the proposed model captures clinically meaningful histopathological patterns that align with known subtype-specific glioma morphology. Importantly, these patterns emerge without requiring pixel-level annotations, demonstrating the capability of the model to implicitly learn prognostically relevant morphological representations from weakly supervised WSI data.

\subsubsection{Importance of Genomic Features}

To further interpret the molecular component of the model, we apply Integrated Gradients (IG) to quantify the contribution of transcriptomic features to the predicted survival risk. The horizontal bar charts in Fig.~\ref{fig:inter} display the top 25 genes ranked by attribution scores. Positive attribution values (red bars) indicate features that increase the predicted hazard risk (poorer prognosis), whereas negative attribution values (blue bars) correspond to features that reduce the predicted risk. As observed in the high-risk GBM patient, genes such as \textit{SPP1} (Secreted Phosphoprotein 1, also known as osteopontin) emerge as strong positive contributors to the predicted risk score. This observation is consistent with previous studies~\cite{wang2025spp1} reporting that \textit{SPP1} is frequently overexpressed in gliomas and is associated with tumor progression, angiogenesis, and unfavorable survival outcomes. In contrast, for the low-risk ODG patient, the model assigns negative attribution values to immune-related genes such as \textit{PRF1} (Perforin 1). PRF1 is a key effector molecule for cytotoxic T cells and natural killer (NK) cells, and its negative risk attribution may reflect transcriptional programs associated with enhanced anti-tumor immune responses. Overall, these attribution patterns suggest that the model captures biologically meaningful molecular signals consistent with known glioma biology.

\subsubsection{Multimodal Synergy}

The joint analysis of pathological and genomic explanations demonstrates that our model leverages truly complementary, mutually reinforcing information. In the high-risk GBM case, the visual encoder pinpoints WSI regions of aggressive tissue remodeling and dense cellularity across multiple scales. Concurrently, the genomic branch highlights the exact molecular underpinnings driving these morphological phenotypes—such as the massive upregulation of \textit{SPP1}, a gene notorious for mediating tumor invasiveness and macrophage recruitment. Conversely, in the ODG case, WSI regions displaying lower-grade oligodendroglial morphology are evaluated alongside a genome-level protective immune signature (\eg, \textit{PRF1}). This synergy, where the visual modality provides the spatial manifestation of tumor burden while the genomic modality uncovers the functional drivers, enables a significantly more robust and deeply interpretable survival prediction compared to unimodal approaches.

\subsubsection{Importance of Global Pan-Cancer Features}

To further investigate the cohort-level biological signals captured by the model, we analyze the importance of global genomic features across five TCGA datasets (as shown in Fig.~\ref{fig:inter}). For each cohort, we aggregate attribution scores to identify the molecular features—spanning both RNA-seq expression and copy number variations (CNVs)—that most strongly influence the predicted survival hazard.

\textbf{Consistency with Local Explanations (GBMLGG):} 
Consistent with the patient-level case studies presented in Sec.~\ref{sec:interpretability}, the global analysis of the GBMLGG cohort identifies \textit{SPP1} as the top-ranked prognostic feature across the entire dataset. This finding aligns with previous studies reporting that \textit{SPP1} is strongly associated with glioma progression, tumor invasion, and poor clinical outcomes. Other highly ranked features, such as \textit{NAMPT} and \textit{CDK4\_cnv}, further reflect key molecular processes underlying glioma biology, including metabolic reprogramming and cell-cycle dysregulation.


\textbf{Multimodal Molecular Signals:} 
The presence of both transcriptomic features (\eg, \textit{SPP1}, \textit{NAMPT}) and genomic alterations (\eg, \textit{CDK4\_cnv}) among the top-ranked predictors illustrates the model’s ability to integrate heterogeneous molecular modalities when estimating survival risk. This multimodal feature prioritization suggests that the model captures complementary biological signals spanning both gene expression dynamics and genomic structural alterations.

\textbf{Cancer-Type–Specific Molecular Patterns:} 
Importantly, the dominant molecular features vary substantially across different cancer cohorts. 
For example, the LUAD cohorts highlight regulators involved in signaling and transcriptional control (\eg, \textit{TRIM26} and \textit{MAP2K2}), while the UCEC cohort prioritizes genes related to epigenetic regulation and cellular signaling pathways (\eg, \textit{ASXL1} and \textit{EXT1}). These observations suggest that the model dynamically adapts its feature attribution to the biological context of each cancer type, rather than relying on a fixed set of global biomarkers.

Overall, the analysis demonstrates that the proposed framework captures biologically meaningful molecular signals across diverse cancer types while maintaining the flexibility to emphasize disease-specific prognostic drivers.

\blue{\subsection{Discussion of Computational Complexity}}

\blue{Our model incorporates multi-scale processing and hierarchical routing, which could otherwise incur substantial computational costs. Direct self-attention over tens of thousands of unaligned patches from three magnifications ($5\times, 10\times, 20\times$) has prohibitive quadratic complexity of $O(M_i^2)$, where $M_i = \sum M_i^s$ denotes the total number of patches for patient $i$. To address this issue, we employ a two-stage sequence reduction strategy (Section 3.3.1). First, a Transformer-based soft-clustering module uses a fixed number of learnable prototype queries ($\mathbf{Q}^s \in \mathbb{R}^{P^s \times D}$), to aggregate $M_i^s$ patch features into $P^s$ semantic prototypes at each scale via a multi-head cross-attention (MHCA). Since $P^s \ll M_i^s$, the aggregation complexity is reduced from $O((M_i^s)^2)$ to $O(P^s \cdot M_i^s)$, which scales linearly with the number of patches. The \textit{Anchor Bridge} then maps these condensed multi-scale prototypes to a compact set of global anchors. This early sequence reduction enables scalable processing of ultra-large WSIs while retaining coarse-to-fine structural information. For hierarchical routing, we replace dense fusion with sparse, multi-level MoE routing. Adaptive Top-$K$ gating activates only relevant experts at each level, facilitating rich cross-modal and multi-scale interactions while maintaining bounded inference costs. To evaluate computational efficiency, we benchmark our model against three representative methods, including MoCAT~\cite{MoCAT_2023_ICCV}, HiMT~\cite{HiMT_2022}, and UMSA~\cite{UMSA_2025}, on a single NVIDIA GeForce RTX 4090 GPU with 24 GB of memory. As reported in Table~\ref{tab:complexity_comparison}, our model contains 12.66M parameters and achieves an inference time of 9.5 ms per WSI, offering a favorable trade-off between computational efficiency and predictive performance.}

\begin{table}[!t]
    \centering
    \renewcommand{\arraystretch}{1.1}
    \setlength\tabcolsep{2.2pt}
    \small
    \caption{\blue{Comparison of computational complexity and efficiency.}}
    \label{tab:complexity_comparison}
    \begin{tabular}{r|cccc}
\toprule
\textbf{Method} & \textbf{\begin{tabular}[c]{@{}c@{}}Params.\\ (M)\end{tabular}} & \textbf{\begin{tabular}[c]{@{}c@{}}FLOPs\\ (G)\end{tabular}}
 & \textbf{\begin{tabular}[c]{@{}c@{}}Peak Memory\\ (MB)\end{tabular}} & \textbf{\begin{tabular}[c]{@{}c@{}}Inference Time\\ (ms/WSI)\end{tabular}} \\ 
\midrule
MoCAT~\cite{MoCAT_2023_ICCV}         & 3.90            & 1.83           & 123.93         & 60.3                \\
HiMT~\cite{HiMT_2022}          & 5.48            & 2.41            & 156.39          & 7.7                  \\
UMSA~\cite{UMSA_2025}          & 31.81           & 40.76            & 553.75          & 72.7                 \\ 
\textbf{Ours}   & 12.66             & 4.99      & 190.16                     & 9.5                       \\ 
  \bottomrule
\end{tabular}
\end{table}


\section{Conclusion}

In this paper, we propose AM$^2$ES, a novel framework for multi-modal cancer survival prediction. AM$^2$ES employs dynamic semantic anchors to mediate the integration of transcriptomic profiles and multi-scale pathological features, thereby facilitating cross-modal alignment. It further incorporates a H-MoE module that decouples phenotypic analysis across tissue scales. Specifically, H-MoE first selects informative features within each scale and then adaptively integrates evidence across scales for each patient. Experiments on multiple cancer cohorts show that AM$^2$ES consistently outperforms state-of-the-art methods while providing fine-grained interpretability of its decision process. Ablation studies further confirm the contribution of each key component.


{
\bibliographystyle{IEEEtran}
\bibliography{pami}
}

\vfill


%





\ifCLASSOPTIONcaptionsoff
  \newpage
\fi

\end{document}